\pdfoutput=1

\documentclass[11pt]{article}

\usepackage[preprint]{acl}

\usepackage{times}
\usepackage{latexsym}

\usepackage[T1]{fontenc}

\usepackage[utf8]{inputenc}

\usepackage{microtype}

\usepackage{inconsolata}

\usepackage{graphicx}

\usepackage{booktabs}
\usepackage{subcaption}
\usepackage{amsmath}

\makeatletter
\newcommand{\anontext}[1]{%
\ifacl@anonymize
[REDACTED FOR ANONYMITY]
\else 
#1
\fi
}
\makeatother

\title{Targeted Multilingual Adaptation for Low-resource Language Families}

\author{
    C.M. Downey$^{\alpha}$ \quad
	Terra Blevins$^{\beta}$ \quad
    Dhwani Serai$^{\alpha}$ \quad
    Dwija Parikh$^{\alpha}$ \quad
    Shane Steinert-Threlkeld$^{\alpha}$\\
    $^\alpha$Department of Linguistics, University of Washington \\
	$^{\beta}$Paul G. Allen School of Computer Science \& Engineering, University of Washington \\
    correspondence: {\tt cmdowney@uw.edu, blvns@cs.washington.edu} \\
}

\begin{document}
\maketitle
\begin{abstract}
The ``massively-multilingual'' training of multilingual models is known to limit their utility in any one language, and they perform particularly poorly on low-resource languages. However, there is evidence that low-resource languages can benefit from \textit{targeted} multilinguality, where the model is trained on closely related languages. To test this approach more rigorously, we systematically study best practices for adapting a pre-trained model to a language family. Focusing on the Uralic family as a test case, we adapt XLM-R under various configurations to model 15 languages; we then evaluate the performance of each experimental setting on two downstream tasks and 11 evaluation languages. Our adapted models significantly outperform mono- and multilingual baselines. Furthermore, a regression analysis of hyperparameter effects reveals that adapted vocabulary size is relatively unimportant for low-resource languages, and that low-resource languages can be aggressively up-sampled during training at little detriment to performance in high-resource languages. These results introduce new best practices for performing language adaptation in a targeted setting.

\end{abstract}

\section{Introduction}\label{sec:intro}

Pre-trained multilingual language models act as the foundation for most current NLP systems outside of English and a few other very high-resource languages. While most languages of the world are relatively data-scarce in comparison to English, multilingual models take the approach of pooling text data across many languages to train a single model that (in theory) covers all training languages \citep[i.a.]{devlin_multilingual_2019, conneau_cross-lingual_2019, conneau-etal-2020-unsupervised, liu-etal-2020-multilingual-denoising, scao_bloom_2023}. In practice, however, massively-multilingual models often perform poorly on low-resource languages \citep{wu-dredze-2020-languages}. 

While multilingual models are susceptible to the so-called ``curse of multilinguality'' --- the observation that overall model performance decreases as more languages are added in pre-training \citep{conneau-etal-2020-unsupervised, wang-etal-2020-negative} --- it is generally accepted that low-resource languages benefit from \textit{some} multilinguality during training, especially when added languages are similar in some way \citep{conneau-etal-2020-unsupervised, ogunremi-etal-2023-mini, chang_when_2023}. Nonetheless, ``massively multilingual'' or ``cross-lingual'' models have remained a central focus of multilingual LLM research \citep[e.g.~][]{ustun_aya_2024}.

This paper joins a growing line of research studying \textit{targeted} multilingualism as a more practical approach to building robust models for mid- and low-resource languages \citep{chang_when_2023, ogueji-etal-2021-small, ogunremi-etal-2023-mini, ljubesic_language_2024}. While studies like \citet{ogunremi-etal-2023-mini} take the approach of training from scratch on a linguistically-informed grouping like a language family, we instead seek to determine the best way to \textit{leverage} existing multilingual models, using their parameters as a starting point for specialization to a more moderate set of languages.

In this work, we systematically evaluate the best technique for adapting a pre-trained multilingual model (XLM-R) to a language family. We use the Uralic family as a case study --- like many families, it includes a few mid-resource languages (e.g.~Hungarian, Finnish) as well endangered and Indigneous languages like Sámi and Erzya, which are extremely data-scarce. Our primary techniques for conducting adaptation are multilingual Language-Adaptive Pre-Training \citep[\textsc{Lapt};][]{chau-etal-2020-parsing} and vocabulary replacement/specialization \citep[i.a.]{dobler-de-melo-2023-focus, downey-etal-2023-embedding}. Our experiments show that both techniques are necessary for robust adapation to the Uralic family.

Importantly, we demonstrate not only is adaptation to a language family as performant or better than training individual models, but also that it is more efficient than monolingual adaptation. We also statistically analyze important factors in multilingual adaptation in order to recommend \textit{best practices} for adapting models to new language families, as measured by down-stream task performance. In particular, we use a regression analysis to assess the impact of \textsc{Lapt} steps, adapted vocabulary size, and language sampling alpha on model performance. Notable results include the fact that specialized vocabularies as small at 16k tokens outperform the cross-lingual XLM-R vocabulary (with 250k tokens), and low-resource languages can be aggressively up-sampled during training without significant degradation of high-resource performance (see \S~\ref{sec:results},\ref{sec:discussion} for more details).

Our contributions are as follows: 1) We train models adapted for the Uralic family that significantly outperform monolingual and multilingual baselines for almost all languages. 2) We conduct a large-scale statistical analysis of important parameters for multilingual adaptation to test their relative effects on downstream task performance. 3) We make best-practice recommendations for adapting cross-lingual models to targeted groupings like language families. 4) We provide an error analysis for Skolt Sámi, which is consistently difficult to model, and discuss the implications and challenges of these results for future work. 5) We make all of our adaptation code, configurations, analysis results, and best-performing Uralic model(s) publicly available at \url{https://github.com/CLMBRs/targeted-xlms}.

\section{Related Work}\label{sec:related_work}

\paragraph{Pre-trained model adaptation}
Extensive work has proposed re-using and modifying pre-trained models for new settings in order to retain existing model knowledge and reduce pre-training costs. \citet{gururangan-etal-2020-dont} show that continued training on domain-specific data effectively adapts pre-trained models to new domains in both high- and low-resource settings. This approach is also used to adapt models to new languages (i.e.~Language-Adaptive Pre-Training / \textsc{Lapt}; \citealp{chau-etal-2020-parsing}).

Other approaches involve training new, language-specific adapter layers to augment a frozen monolingual \cite{artetxe-etal-2020-cross} or multilingual encoder \cite{pfeiffer-etal-2020-mad, ustun-etal-2020-udapter, faisal-anastasopoulos-2022-phylogeny}. A comparison of these cross-lingual adaptation approaches \cite{ebrahimi-kann-2021-adapt} found that continued pre-training often outperforms more complex setups, even in low-resource settings.

\citet{acs-etal-2021-evaluating} investigate the transferability of monolingual BERT models for Uralic languages specifically. They find that vocabulary overlap and coverage is extremely important for transfer success, and also that the importance of language-relatedness is questionable, since English and Russian BERT transfer well to Uralic languages written in Latin and Cyrillic script, respectively.

\paragraph{Model vocabulary and script}
A major limitation to adapting pre-trained models to new languages is the subword vocabulary, which often fails to cover unseen scripts \cite{pfeiffer-etal-2021-unks} or tokenizes target text inefficiently \cite{acs_exploring_2019, ahia_all_2023}. \citet{muller-etal-2021-unseen} demonstrate that script is another important factor in predicting transfer success: pre-trained coverage of closely-related languages improves transfer, but only if the target language is written in the same script as its pre-trained relative.

A range of adaptation techniques have been proposed to overcome this tokenization issue, such as extending the vocabulary with new tokens \citep{chau-etal-2020-parsing, wang-etal-2020-extending, liang-etal-2023-xlm} or completely replacing and re-training the vocabulary and embedding matrix from a random initialization \citep{artetxe-etal-2020-cross, de-vries-nissim-2021-good}. Other work reuses information in pre-trained embeddings rather than initializing new ones at random. This may include scaling up smaller embedding spaces from models trained on the target language \cite{de-vries-nissim-2021-good, ostendorff_efficient_2023} or copying embeddings from the original vocabulary where there is exact vocabulary overlap \cite{pfeiffer-etal-2021-unks}.

In this study, we follow a line of recent work that re-initializes vocabulary and embeddings based on the structure of the embedding space for the original model \citep[i.a.]{minixhofer-etal-2022-wechsel, ostendorff_efficient_2023}. \citet{dobler-de-melo-2023-focus} introduce the \textsc{Focus} algorithm, which like \citet{pfeiffer-etal-2021-unks} carries over original embeddings where there is an exact match with the new vocabulary. For new tokens however, it initializes embeddings as a linear combination of the old embeddings for the most semantically similar tokens, as computed by an auxiliary embedding model. As an alternative, \citet{downey-etal-2023-embedding} propose three simple heuristics for initializing a new embedding matrix, one being the familiar strategy of carrying over the embeddings of overlapping tokens, and the others involving initializing new tokens based on script-wise distributions in the original space. They compare these methods to the \textsc{Focus} algorithm and find the latter has only a small advantage over the heuristic-based techniques.

\paragraph{Targeted multilingualism}
A recent line of work has proposed models trained with \textit{targeted} or \textit{linguistically-informed} multilingualism, as opposed to the ``massively-multilingual'' approach covering as many languages as feasible \citep[e.g.~][]{conneau-etal-2020-unsupervised, scao_bloom_2023}. Notably, \citet{chang_when_2023} show that while massively-multilingual models hurt individual language performance, low-resource languages in particular benefit from \textit{limited} multilinguality, especially when the added languages are syntactically similar (e.g.~similar word order).

Examples of targeted multilingual approaches include \citet{ogueji-etal-2021-small}, who train a multilingual model from scratch on 11 African languages and show performance that is as good or better than XLM-R. \citet{ogunremi-etal-2023-mini} refine this approach by showing that multilingual training on languages from individual African language families is more data-efficient than using a mixture of unrelated African languages. \citet{snaebjarnarson-etal-2023-transfer} also show success for the low-resource language Faroese by training a multilingual model on its close Germanic relatives.

Other work investigates using multilingual training with related languages as an \textit{adaptation} process, starting from a pre-trained cross-lingual model rather than training from scratch. \citet{alabi-etal-2022-adapting} adapt XLM-R to the 17 highest-resource African languages via \textsc{Lapt}, while also removing XLM-R vocabulary items that are unused for the target languages. \citet{ljubesic_language_2024} use \textsc{Lapt} to adapt XLM-R to the very closely related Slavic languages of Bosnian, Croatian, Montenegrin, and Serbian. \citet{senel-etal-2024-kardes} adapt XLM-R separately to five low-resource Turkic languages, showing that including the high-resource Turkish language during training improves this adaptation.

The present work systematically analyzes which factors are responsible for the success of targeted multilingual adaptation. We focus on the model adaptation paradigm since cross-lingual models learn useful language-general patterns that can be leveraged for a ``warm-start'' to training \citep{conneau-etal-2020-emerging}. Unlike \citet{ljubesic_language_2024, senel-etal-2024-kardes}, we specialize model vocabulary for the target language(s), since cross-lingual tokenizers typically perform poorly for low-resource languages \citep{rust-etal-2021-good}. We follow \citet{dobler-de-melo-2023-focus} and \citet{downey-etal-2023-embedding} in using a vocabulary specialization technique that leverages the structure of the original model embedding space, while creating a new vocabulary that is directly optimized for the target languages, in contrast to \citet{alabi-etal-2022-adapting}, which simply uses a subset of the original model vocabulary. Finally, we follow \citet{ogunremi-etal-2023-mini} in conducting adaptation for a language family, while keeping in mind the observation from \citet{senel-etal-2024-kardes} that including a high-resource language during adaptation can be advantageous. This comes naturally with our chosen testbed of the Uralic family, which contains both high- and low-resource languages.

\section{Experiments}
\label{sec:experiments}
Our experiments are designed to assess the best method for adapting a pre-trained cross-lingual model to a specific language family (in our case, Uralic). We are especially interested in identifying conditions that produce the best model(s) for low-resource family members. Our primary approach employs Language-Adaptive Pre-Training \citep[\textsc{Lapt},][]{chau-etal-2020-parsing} on a dataset of Uralic languages, as well as vocabulary specialization \citep[i.a.]{downey-etal-2023-embedding}. Adapted models are compared to both multilingual and monolingual baselines.

Within our multilingual experiments, we search a range of important hyper-parameters and explicitly model their influence on downstream performance using a linear mixed-effects regression. Namely, we test the effect of number of \textsc{Lapt} steps, size of the language-specialized vocabulary, and the $\alpha$ parameter controlling multinomial language sampling distribution during \textsc{Lapt} \citep{conneau_cross-lingual_2019, conneau-etal-2020-unsupervised}.

\begin{table*}[h]
    \centering
    \resizebox{\textwidth}{!}{
    \begin{tabular}{lccccccc}
        \toprule
        Language & Code & Branch & Script & XLM-R Data (GB) & \textsc{Lapt} Data (GB) & \textsc{Lapt} Data (lines) & Sources \\
        \midrule
        Russian & ru & n/a & Cyrillic & 278.0 & 9.1 & $32.7 \times 10^6$ & O \\
        Hungarian & hu & Hungarian & Latin & 58.4 & 12.8 & $64.8 \times 10^6$ & O \\
        Finnish & fi & Finnic & Latin & 54.3 & 9.3 & $50.2 \times 10^6$ & O \\
        Estonian & et & Finnic & Latin & 6.1 & 2.8 & $15.8 \times 10^6$ & O \\
        Komi & koi & Permic & Cyrillic & 0 & $6.8 \times 10^{-3}$ & $48.5 \times 10^3$ & OPJ \\
        Mari & mhr/mrj & Mari & Cyrillic & 0 & $6.5 \times 10^{-3}$ & $25.3 \times 10^3$ & OJ \\
        Erzya & myv & Mordvinic & Cyrillic & 0 & $6.0 \times 10^{-3}$ & $32.6 \times 10^3$ & PJ \\
        Veps & vep & Finnic & Latin & 0 & $5.3 \times 10^{-3}$ & $35.7 \times 10^3$ & P \\
        Udmurt & udm & Permic & Cyrillic & 0 & $4.3 \times 10^{-3}$ & $28.1 \times 10^3$ & PJ \\
        Sámi & se/sme & Sámi & Latin & 0 & $3.9 \times 10^{-3}$ & $34.5 \times 10^3$ & PJ \\
        Karelian & krl & Finnic & Latin & 0 & $2.4 \times 10^{-3}$ & $17.4 \times 10^3$ & PJ \\
        Moksha & mdf & Mordvinic & Cyrillic & 0 & $1.2 \times 10^{-3}$ & $9.3 \times 10^3$ & P \\
        Livonian & liv & Finnic & Latin & 0 & $0.5 \times 10^{-3}$ & $14.2 \times 10^3$ & P \\
        Votic & vot & Finnic & Latin & 0 & $<0.1 \times 10^{-3}$ & 474 & P \\
        Ingrian & izh & Finnic & Latin & 0 & $<0.1 \times 10^{-3}$ & 21 & P \\
        \bottomrule
    \end{tabular}}
    \caption{Listing of available training data by language (after cleaning, de-duplicating, and reserving 10\% for eval and test sets). XLM-R data is the amount of data used to pre-train that model. \textsc{Lapt} data is the amount of data available for adaptive training on Uralic languages in our experiments. Codes for language data sources: O = OSCAR, P = OPUS, J = JHUBC.}
    \label{tab:data}
\end{table*}

\paragraph{Languages}
The first step of our adaptation process is to obtain raw-text \textsc{Lapt} data for as many Uralic languages as possible. For the high-resource languages (Estonian, Finnish, Hungarian, and Russian), we obtain all training data from the multilingual OSCAR corpus v.22.01 \citep{abadji-etal-2022-towards}. This corpus also contains a small amount of raw text for the low-resource languages Komi (\texttt{koi}) and Mari (\texttt{mhr/mrj}). We further source low-resource language data from monolingual splits of the OPUS translation corpus \citep{tiedemann-nygaard-2004-opus} and the Johns Hopkins University Bible Corpus \citep{mccarthy-etal-2020-johns}.

An inventory of \textsc{Lapt} text data is found in Table~\ref{tab:data}. This represents the total amount of data after combining all corpora for each language. We cover 6/8 Uralic branches, lacking only Ob-Ugric and Samoyedic \citep{austerlitz_uralic_2008}. The resource gap between the high- and low-resource languages is stark: Estonian (the fourth-highest-resource language) has approximately 1000x more data than the next highest (Komi). These four highest-resource languages were also included in the training data for XLM-R, while the remainder were not. We treat this as the cutoff point between the ``high-resource'' and ``low-resource'' Uralic languages for the remainder of this work.

We include Russian as a high-resource language, though it is not Uralic. Many Uralic languages are spoken by ethnic minorities within Russia and the former Soviet Union, and use modified forms of the Russian Cyrillic alphabet. The lack of a high-resource Uralic language written in Cyrillic could be a problem for low-resource language performance, since script overlap has been shown to be a vital ingredient in cross-lingual transfer \citep{muller-etal-2021-unseen, downey-etal-2023-embedding}. Further, Russian is a major source of loan-words for Uralic languages, as well as an official language throughout Russian territory \citep{austerlitz_uralic_2008}.

\begin{figure}[h]
    \begin{center}
    \includegraphics[width=0.48\textwidth]{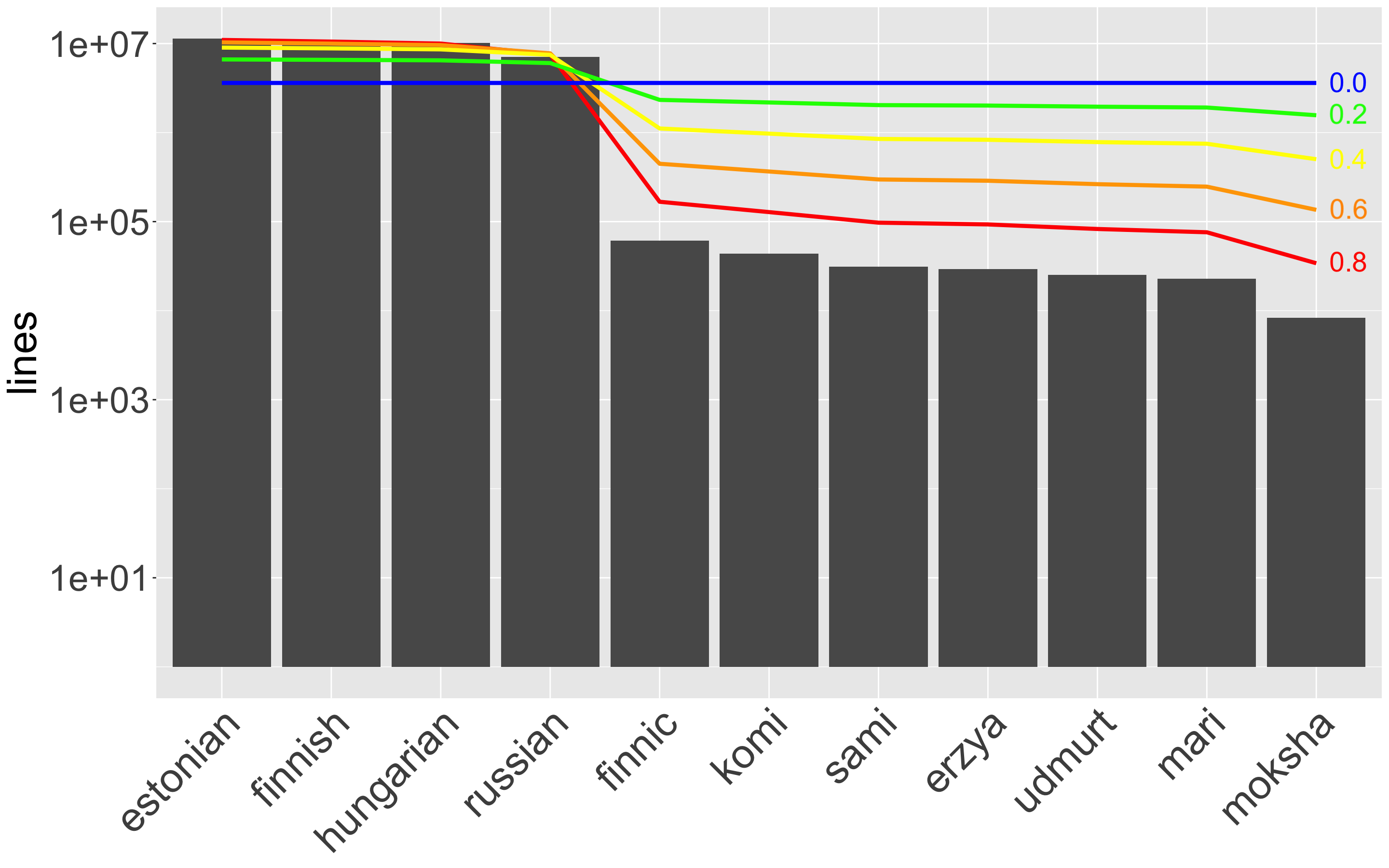}
    \caption{Uralic data composition by number of lines, on a log scale. The actual data quantities are shown with bars, while sampling distributions with several values of the $\alpha$ parameter are plotted as lines}
    \label{fig:language_composition}
    \end{center}
\end{figure}

During our experiments, we sample languages according to a multinomial distribution parameterized by the hyper-parameter $\alpha$ (\citealp[i.a.]{conneau_cross-lingual_2019,conneau-etal-2020-unsupervised}; see Figure~\ref{fig:language_composition}). Languages are sampled sentence-wise rather than batch-wise, meaning multiple languages can be sampled in each batch. 

\paragraph{Vocabulary replacement}
To specialize the model's vocabulary for target languages, we first train a new Sentencepiece model \citep{kudo-richardson-2018-sentencepiece} on 5 million lines sampled from the training set.\footnote{When adapting to single languages with < 5 million lines, the vocabulary is trained on the entire training set.} For simplicity, we train multilingual tokenizers with a consistent sampling parameter of $\alpha=0.2$.\footnote{Pilot experiments suggest the choice of $\alpha$ during vocabulary initialization is not as important as the value picked during multilingual training.} Once a new vocabulary is formed, we re-initialize the model's embedding matrix using the \textsc{Focus} algorithm introduced by \citet{dobler-de-melo-2023-focus}. We test the effect of vocabulary size by training specialized vocabularies with 16k, 32k, and 64k tokens.\footnote{Throughout this paper, 16k, 32k, and 64k are shorthand for $2^{14}$, $2^{15}$, and $2^{16}$ respectively.}

\paragraph{Training}

All experiments use XLM-R base as a starting point \citep{conneau-etal-2020-unsupervised}. We conduct \textsc{Lapt} on the multilingual Uralic dataset for 100k, 200k, or 400k steps. Following \citet{downey-etal-2023-embedding}, for experiments with vocabulary specialization, the transformer blocks are frozen for the first 10k steps, then unfrozen for the remainder, to prevent model overfitting on the initial (possibly poor) embedding initializations. The checkpoint with the best MLM loss on a development set is selected for task fine-tuning and evaluation.

For our shortest experiments (100k steps) we test four values of $\alpha$: $\{0.1, 0.2, 0.3, 0.4\}$. For longer experiments, we test only the two most promising values: $\{0.1, 0.2\}$. Because the data ratio between our high and low-resource languages is so extreme (Table~\ref{tab:data}), we cap the four high-resource languages at approximately 2 GB of text each.\footnote{This is in addition to alpha sampling, reflected in Figure~\ref{fig:language_composition}.} Because several languages of the Finnic branch have less than 1 MB of text, we also sample the 5 low-resource Finnic languages as if they are a single language (``Finnic'' in Figure~\ref{fig:language_composition}). This is to prevent extreme over-sampling of tiny datasets such as Ingrian. 

\paragraph{Task evaluation}
We evaluate model performance with Part Of Speech (POS) tagging accuracy as well as Unlabeled Attachment Score (UAS), a metric for syntactic dependency parsing. Both of these evaluations are conducted on Universal Dependencies (UD) treebanks \cite{de-marneffe-etal-2021-universal}.\footnote{Currently, UD appears to be the only source for high-quality NLP evaluation data in low-resource Uralic languages.} Treebanks are available for all high-resource languages plus Erzya, North Sámi (\texttt{sme}), Komi, Karelian, Livvi, Moksha, and Skolt Sámi (\texttt{sms}). Models are fine-tuned for each task over four random seeds.

Because the available amount of fine-tuning data varies considerably over languages, we consider three evaluation settings: \textit{few-shot}, \textit{full-finetune}, and \textit{zero-shot}. In the \textit{few-shot} setting, models are fine-tuned on 512 sampled sentences per language. For \textit{full-finetune}, models are fine-tuned on the entirety of the fine-tuning data for each language (ranging from 896 sentences for Erzya to 32,768 for Russian). We additionally employ the \textit{zero-shot} setting because, with the exception of Erzya and North Sámi, the low-resource languages we consider only have small test sets, with no standard training data. For this setting, we fine-tune the model on the full collection of languages with training sets, and then evaluate directly on the target test set. An inventory of Uralic UD evaluation data can be found in Table~\ref{tab:eval_data_inventory} of Appendix~\ref{app:evaluation}, along with more details on our evaluation methodology.

\paragraph{Baselines}
Our simplest baseline is ``off-the-shelf'' XLM-R --- the pre-trained model from \citet{conneau-etal-2020-unsupervised} with no modifications. We also test XLM-R adapted with \textsc{Lapt}, but without vocabulary specialization. \textsc{Lapt} alone is a strong baseline. However, as \citet{downey-etal-2023-embedding} note, keeping a large ``cross-lingual'' vocabulary during \textsc{Lapt} incurs considerable extra computational cost compared to training a smaller, specialized vocabulary. Given the observation that cross-lingual tokenizers are inefficient and ineffective for low-resource languages \citep{acs_exploring_2019, rust-etal-2021-good}, we hypothesize a specialized vocabulary will show a performance advantage in addition to the reduction in computational cost.

We also compare our multilingual models to baselines adapted to single languages. While multilingualism is known to help low-resource languages to some degree, it is also an open question in what circumstances multilingualism becomes a ``curse'' \citep{conneau-etal-2020-unsupervised, chang_when_2023}. To make this comparison, we adapt XLM-R with \textsc{Lapt} on individual languages, with a vocab size of 16k per language, and assuming a shared computational ``budget'' of 400k training steps. The steps are allocated across languages according to the multinomial distribution with $\alpha = 0.1$ , similar to the data sampling technique for multilingual training. We thus design this baseline to be roughly comparable to our multilingual model trained with 400k steps, vocab size 16k, and $\alpha = 0.1$.

\section{Results}\label{sec:results}

\begin{table*}[h]
    \centering
    \resizebox{\textwidth}{!}{
    \begin{tabular}{ccccccccc}
        \toprule
        Task & Type & Erzya & North Sámi & Estonian & Finnish & Hungarian & Russian & Avg \\
        \midrule
        UAS & monolingual & 49.7 $\pm$ 0.7 & 42.0 $\pm$ 2.2 & 52.4 $\pm$ 1.0 & \textbf{69.2 $\pm$ 2.1} & 63.2 $\pm$ 3.4 & 69.1 $\pm$ 1.8 & 57.6 \\
        UAS & multilingual & \textbf{58.8 $\pm$ 2.3} & \textbf{51.3 $\pm$ 0.5} & \textbf{56.9 $\pm$ 2.5} & \textbf{71.2 $\pm$ 2.1} & \textbf{69.9 $\pm$ 1.2} & \textbf{71.7 $\pm$ 2.6} & \textbf{63.3} \\
        \midrule
        POS & monolingual & 62.0 $\pm$ 1.3 & 60.8 $\pm$ 2.0 & \textbf{84.0 $\pm$ 0.6} & \textbf{79.1 $\pm$ 2.3} & 85.9 $\pm$ 2.2 & 86.5 $\pm$ 1.8 & 76.4 \\
        POS & multilingual & \textbf{76.1 $\pm$ 3.3} & \textbf{73.2 $\pm$ 1.2} & 77.7 $\pm$ 3.9 & \textbf{79.7 $\pm$ 2.6} & \textbf{89.3 $\pm$ 1.3} & \textbf{87.5 $\pm$ 0.5} & \textbf{80.6} \\
        \bottomrule
    \end{tabular}}
    \caption{Few-shot comparisons with monolingual baselines (both tasks). All models have vocabulary size 16k. Multilingual models are trained for 400k steps with $\alpha=0.1$. Monolingual models trained for a total of 400k steps ``budgeted'' across the languages, according to $\alpha=0.1$, as described in \S~\ref{sec:experiments}.}
    \label{tab:fewshot_mono_baseline}
\end{table*}

\begin{table*}[h]
    \centering
    \begin{tabular}{cccccccc}
        \toprule
        Task & Type & Karelian & Komi & Livvi & Moksha & Skolt Sámi & Avg \\
        \midrule
        UAS & monolingual & 61.7 $\pm$ 0.4 & 28.4 $\pm$ 4.6 & 61.1 $\pm$ 0.8 & 40.0 $\pm$ 3.1 & 28.9 $\pm$ 2.1 & 44.0 \\
        UAS & multilingual & \textbf{65.9 $\pm$ 0.3} & \textbf{73.8 $\pm$ 0.6} & \textbf{65.9 $\pm$ 0.3} & \textbf{70.2 $\pm$ 0.2} & \textbf{41.4 $\pm$ 1.6} & \textbf{63.4} \\
        \midrule
        POS & monolingual & 84.5 $\pm$ 0.1 & 44.6 $\pm$ 3.1 & 81.6 $\pm$ 0.2 & 49.7 $\pm$ 2.0 & 52.6 $\pm$ 0.5 & 62.6 \\
        POS & multilingual & \textbf{87.7 $\pm$ 0.2} & \textbf{80.1 $\pm$ 0.3} & \textbf{85.0 $\pm$ 0.2} & \textbf{78.3 $\pm$ 0.2} & \textbf{55.4 $\pm$ 0.3} & \textbf{77.3} \\
        \bottomrule
    \end{tabular}
    \caption{Zero-shot comparisons with monolingual baselines (both tasks) with the same models as Table \ref{tab:fewshot_mono_baseline}. Monolingual models are fine-tuned on the most similar language with a UD training set: Finnish $\rightarrow$ Karelian, Livvi; Erzya $\rightarrow$ Komi, Moksha; North Sámi $\rightarrow$ Skolt Sámi.}
    \label{tab:zeroshot_mono_baseline}
\end{table*}

\begin{table*}[h]
    \centering
    \resizebox{\textwidth}{!}{
    \begin{tabular}{cccccccccc}
        \toprule
        \textsc{Lapt} & Alpha & Vocab & Erzya & North Sámi & Estonian & Finnish & Hungarian & Russian & Avg \\
        \midrule
        0 & * & 250k (orig) & 29.0 $\pm$ 2.1 & 26.2 $\pm$ 1.0 & 37.4 $\pm$ 5.4 & 51.5 $\pm$ 3.1 & 45.3 $\pm$ 10.0 & 47.6 $\pm$ 3.5 & 39.5 \\
        400k & 0.1 & 250k (orig) & 54.0 $\pm$ 0.9 & 51.0 $\pm$ 1.3 & 54.7 $\pm$ 2.3 & 71.2 $\pm$ 1.0 & 69.1 $\pm$ 1.4 & 70.1 $\pm$ 3.4 & 61.7 \\
        \midrule
        400k & 0.1 & 16k & 58.8 $\pm$ 2.3 & 51.3 $\pm$ 0.5 & 56.9 $\pm$ 2.5 & 71.2 $\pm$ 2.1 & 69.9 $\pm$ 1.2 & 71.7 $\pm$ 2.6 & 63.3 \\
        400k & 0.1 & 32k & 56.6 $\pm$ 0.8 & 52.0 $\pm$ 0.8 & 56.7 $\pm$ 1.9 & 72.0 $\pm$ 1.8 & 70.1 $\pm$ 0.8 & 71.9 $\pm$ 2.0 & 63.2 \\
        400k & 0.1 & 64k & \textbf{61.5 $\pm$ 2.8} & \textbf{53.8 $\pm$ 0.8} & \textbf{60.7 $\pm$ 0.9} & \textbf{73.0 $\pm$ 1.0} & \textbf{75.2 $\pm$ 0.5} & \textbf{74.2 $\pm$ 2.2} & \textbf{66.4} \\
        \bottomrule
    \end{tabular}}
    \caption{Few-shot UAS --- comparison with multilingual baselines. First row is XLM-R ``off-the-shelf'' (without \textsc{Lapt} or vocabulary specialization). Second row is XLM-R with original cross-lingual vocabulary, but fine-tuned on Uralic languages with \textsc{Lapt}}
    \label{tab:fewshot_uas_baselines}
\end{table*}

\begin{figure*}[h]
    \begin{center}
    \includegraphics[width=\textwidth]{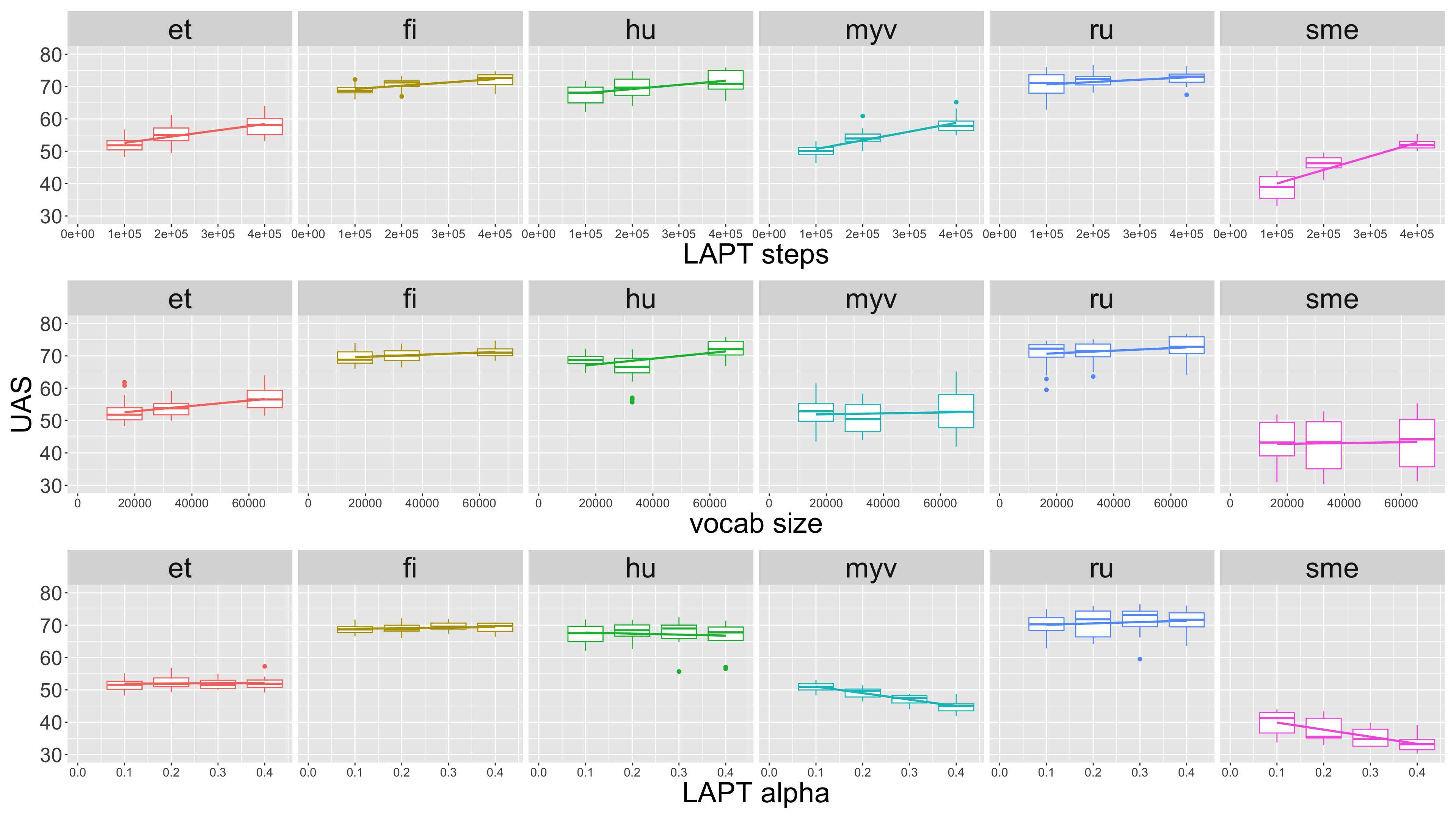}
    \caption{Few-shot UAS --- effect of hyper-parameters by language, marginalized across other parameter settings}
    \label{fig:fewshot_uas_plots}
    \end{center}
\end{figure*}

\begin{table*}[h]
    \centering
    \resizebox{\textwidth}{!}{
    \begin{tabular}{ccccccccc}
        \toprule
        \textsc{Lapt} & Alpha & Vocab & Karelian & Komi & Livvi & Moksha & Skolt Sámi & Avg \\
        \midrule
        0 & * & 250k (orig) & 59.0 $\pm$ 0.4 & 41.1 $\pm$ 1.4 & 56.0 $\pm$ 0.9 & 52.7 $\pm$ 0.03 & 44.4 $\pm$ 1.4 & 50.6 \\
        400k & 0.1 & 250k (orig) & 65.2 $\pm$ 0.3 & 73.9 $\pm$ 0.4 & 63.4 $\pm$ 0.4 & 70.4 $\pm$ 0.6 & \textbf{44.8 $\pm$ 1.2} & 63.6 \\
        \midrule
        400k & 0.1 & 16k & 65.9 $\pm$ 0.3 & 73.8 $\pm$ 0.6 & \textbf{65.9 $\pm$ 0.2} & 70.2 $\pm$ 0.2 & 41.4 $\pm$ 1.6 & 63.4 \\
        400k & 0.1 & 32k & \textbf{66.4 $\pm$ 0.4} & 74.9 $\pm$ 0.3 & 65.4 $\pm$ 0.7 & 71.7 $\pm$ 0.7 & 43.3 $\pm$ 1.5 & \textbf{64.3} \\
        400k & 0.1 & 64k & 66.0 $\pm$ 0.4 & \textbf{75.0 $\pm$ 0.1} & 65.6 $\pm$ 0.5 & \textbf{73.3 $\pm$ 0.5} & 40.8 $\pm$ 1.3 & 64.1 \\
        \bottomrule
    \end{tabular}}
    \caption{Zero-shot UAS --- comparison with multilingual baselines. First row is XLM-R ``off-the-shelf'' (without \textsc{Lapt} or vocabulary specialization). Second row is XLM-R with original cross-lingual vocabulary, but fine-tuned on Uralic languages with \textsc{Lapt}}
    \label{tab:zeroshot_uas_baselines}
\end{table*}

\begin{figure*}[h]
    \begin{center}
    \includegraphics[width=0.85\textwidth]{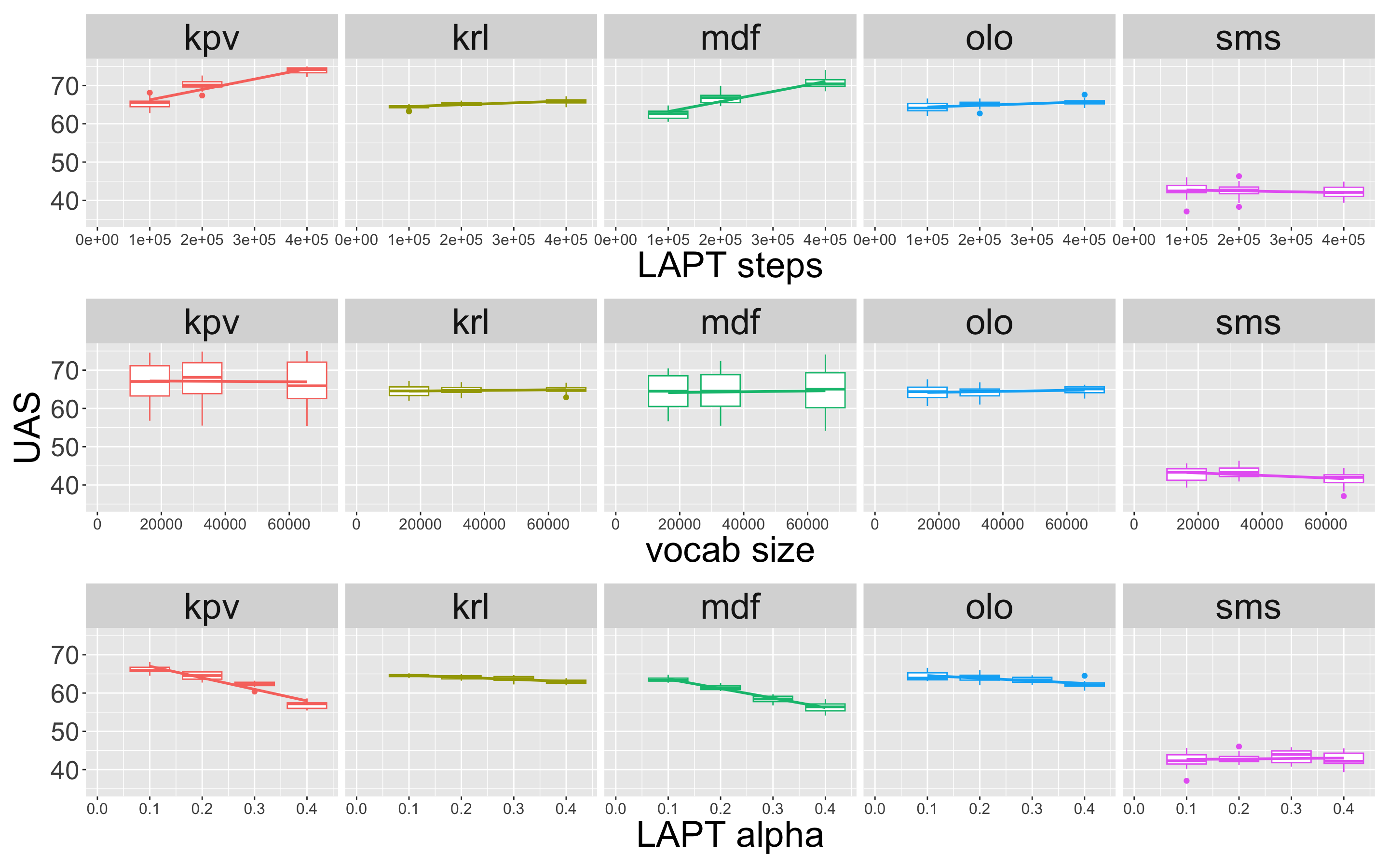}
    \caption{Zero-shot UAS --- effect of hyper-parameters by language, marginalized across other parameter settings}
    \label{fig:zeroshot_uas_plots}
    \end{center}
\end{figure*}

We present our results in two main sections. First, we compare our best-performing Uralic-adapted multilingual models to both multilingual and monolingual baselines. We show that our chosen method of layering \textsc{Lapt} and vocabulary specialization on a pre-trained multilingual model largely outperforms alternatives on downstream tasks and is more computationally efficient. 

We then analyze the dynamics of important factors during multilingual adaptation such as number of \textsc{Lapt} steps, adapted vocabulary size, and sampling alpha. Our grid search of hyper-parameters for multilingual \textsc{Lapt} yields 72 evaluation data-points per language, per task, per setting.\footnote{3 training lengths $\times$ 3 vocabulary sizes $\times$ 2 alpha values $\times$ 4 random seeds (during fine-tuning) = 72. Only 2 alpha values are tested over all training lengths.} We first visualize and discuss the overall trends observed for each parameter; then, we present a regression analysis of the combined effect of these parameters on task performance.

\subsection{Baselines}\label{sec:baseline_results}

\paragraph{Monolingual baselines}
Tables~\ref{tab:fewshot_mono_baseline} and \ref{tab:zeroshot_mono_baseline} compare our best-performing, fully-adapted multilingual models to the comparable monolingual baselines described in \S\ref{sec:experiments}. With a few exceptions for high-resource languages like Estonian and Finnish, the multilingual models substantially outperform the baselines. This is especially salient for the UAS task (first two rows of each table), the \textit{zero-shot} setting (Table~\ref{tab:zeroshot_mono_baseline}), and low-resource languages.

\paragraph{Multilingual baselines}
Tables~\ref{tab:fewshot_uas_baselines} and \ref{tab:zeroshot_uas_baselines} show a comparison of our fully-adapted multilingual models to multilingual baselines for the dependency parsing task. The first row in each represents XLM-R ``off-the-shelf'' --- the original model without \textsc{Lapt} or adjustments to the vocabulary. The second row is the XLM-R adapted with \textsc{Lapt}, but without vocabulary specialization. It retains the large ``cross-lingual'' vocabulary inherited from XLM-R, which is almost 4x larger than our largest adapted vocabulary (64k tokens). 

Table~\ref{tab:fewshot_uas_baselines} shows that in \textit{few-shot} evaluations, our smallest model with vocabulary specialization significantly outperforms the best baseline model without. Creating an adapted vocabulary of 16k tokens results in an average performance gain of 1.6 over the baseline, and increasing to 64k tokens yields an improvement of 4.7 points. We also note that conducting \textsc{Lapt} on XLM-R with its original vocabulary incurs approximately 2-3x more computational cost than training a version with a specialized vocabulary of size 32k \citep{downey-etal-2023-embedding}.

In contrast, the \textit{zero-shot} evaluations do not reflect this consistent improvement with increasing adapted vocabulary size (Table~\ref{tab:zeroshot_uas_baselines}; this is also reflected in our statistical analysis later in this section). 4 of the 5 zero-shot languages still see their best results when modeled with a specialized vocabulary. The exception is Skolt Sámi, which is modeled best by the +\textsc{Lapt}/-vocab-adaptation baseline. However, as we will note several times, our results for Skolt Sámi go against overall trends in our experiments, and we delve into this finding further with an error analysis in \S\ref{sec:discussion}.

For space and clarity, we have focused only on the UAS results in this section. The comparable tables for POS can be found in Appendix~\ref{app:results}. For POS, we observe similar trends to UAS, though the \textsc{Lapt} baseline with the original vocabulary is more on par with the specialized vocabulary settings. We hypothesize that this is reflective of POS tagging being an overall simpler task than dependency parsing, since the latter requires more advanced knowledge of linguistic structure. We believe it is telling, therefore, that the advantage of specialized-vocabulary models is clearer in the more complicated UAS task.

\subsection{Qualitative trends}\label{sec:qualitative_trends}

Figure~\ref{fig:fewshot_uas_plots} shows visualizations of the per-language effect of each hyper-parameter (marginalized across other parameters) in the \textit{few-shot} setting. These plots show the UAS experiments, but they reflect overall trends seen in our statistical data analysis across both tasks.\footnote{A corresponding visualization for POS can be found in Figure~\ref{fig:fewshot_pos_plots} in the Appendix.} First, the number of \textsc{Lapt} (training) steps unsurprisingly has a large effect on performance across languages; this reflects that the adapted model may take a long time to properly converge on new languages. This may be supported by the slope being steeper for languages that are new to XLM-R such as Erzya (myv). Second, adapted vocabulary size seems to have an overall positive effect on performance. However, this effect is not as strong as adding more \textsc{Lapt} steps and not as clear for the low-resource languages Erzya (myv) and North Sámi (sme). Finally, the effect of sampling alpha diverges between high- and low-resource languages, as lower alpha values up-sample low-resource languages and down-sample high-resource ones. More notable is the fact that the performance gain for low-resource languages at lower alpha values is much greater than the corresponding degradation on high-resource languages.

Equivalent plots for the \textit{zero-shot} setting are found in Figure~\ref{fig:zeroshot_uas_plots}. The effects of training steps and alpha are similar to the \textit{few-shot} trends. However, the choice of vocabulary size does not have an obvious effect in this setting, an observation that is corroborated by our statistical analysis. Also of note is the fact that the performance for Skolt Sámi remains consistently poor across hyperparameters, which we investigate further in \S~\ref{sec:discussion}.

\subsection{Statistical analysis}\label{sec:statistics}

\paragraph{Experimental Setup}
We conduct our regression analysis with linear mixed-effect models in the \texttt{lme4} package for R \citep{bates_fitting_2015}. We predict model performance with the following predictors: \textsc{Lapt} steps and vocabulary size as fixed continuous effects; for the \textit{few-shot} and \textit{full-finetune} settings, fine-tuning examples are also modeled this way; we include task (POS vs UAS) as a fixed categorical effect, following the observation that results for the two tasks mostly mirror each other, modulo a fixed offset (POS accuracy is higher than UAS). We justify this by testing a version of the regression with interaction terms between the task and other hyper-parameters (e.g.~steps), but find no significant interactions. ANOVA confirms no significant difference from the model without task-interactions ($p=0.95$).

Because the effect of $\alpha$ shows a different sign and magnitude between high- and low-resource languages, we model it as an \textit{interaction} with a binary categorical variable representing whether the language is high- or low-resource. We justify the binary variable by the stark jump in resources between these two categories (see Section~\ref{sec:experiments}). 

Finally, because of the complex factors leading to differing baseline performance between languages, we include a language-wise random-effect intercept. The final formula for this regression, as well as the full summary table with coefficients, can be found in the Appendix, Table~\ref{tab:finetuned_regression_a}.

\paragraph{Few-shot / Full-finetune Results}
We find highly significant effects on performance ($p < 0.001$) for \textsc{Lapt} steps, vocabulary size, fine-tuning examples, and task.\footnote{Effect of task simply means baseline scores of each are different --- about 14 points lower for UAS.} Sampling alpha is significant in the low-resource case ($p = 0.035$), but not for high-resource languages ($p = 0.36$). This indicates choosing a lower alpha has a significant positive effect for low-resource language performance, without significantly hurting high-resource performance. The coefficient estimate for steps is 1.67, meaning an overall gain of 1.67 POS/UAS points for each 100k steps. The estimate for vocabulary size is 0.62 points per 16k tokens. The estimate for fine-tuning examples is 0.40 per 512 examples. In terms of our experiments, this means that doubling the number of steps from 100k to 200k is $\sim2.7$ times as effective as doubling the vocabulary from 16k to 32k, and $\sim4.2$ times as effective as doubling the number of fine-tuning examples to 1024. The estimate for alpha in the low-resource case is $-1.36$, meaning performance for low-resource languages drops about that much when alpha is raised from 0.1 to 0.2. Finally, we also test for, but find no significant interaction between, steps and vocabulary size; we confirm with ANOVA comparison that there is no significant difference between models with and without this interaction ($p=0.43$). 

\paragraph{Zero-shot Results}
Our regression for the \textit{zero-shot} setting is similar to the previous, except that there is no variable for number of fine-tuning examples (which is not applicable for zero-shot transfer), and there is no interaction between sampling alpha and resource level, since all considered zero-shot languages are low-resource. The effects for steps and task are highly significant ($p < 0.001$); alpha is also significant ($p = 0.0027$). In contrast to the fine-tuned settings, vocabulary size is not significant ($p = 0.73$). The estimate for steps is 1.35 points per 100k steps. The estimate for alpha is $-0.81$ per increment of 0.1. These estimates are slightly smaller in magnitude than for the \textit{few-shot}/\textit{full-train} experiments; this could be partly due to the results for Skolt Sámi, which shows little change under any hyper-parameter configuration.

\section{Discussion}\label{sec:discussion}
Our discussion will first address the consistently poor performance seen on Skolt Sámi tasks (sms, \S\ref{sec:sami}). After this, we will move to the best practices suggested by our experimental results (\S\ref{sec:takeaways}).

\subsection{Skolt Sámi error analysis}\label{sec:sami}
The consistently poor Skolt Sámi task performance across experimental settings suggests that the Sámi \textsc{Lapt} data may not be useful for this variant. We note that the datasets used for \textsc{Lapt} (in the case of Sámi, OPUS \cite{tiedemann-nygaard-2004-opus} and the JHUBC \cite{mccarthy-etal-2020-johns}) label most text as either undifferentiated Sámi (\texttt{se}) or as North Sámi (\texttt{sme}); however, Sámi is a group of languages, not all of which are mutually intelligible.

We therefore consider multiple tests for distribution shifts between the \textsc{Lapt} data and UD evaluation. The first is tokenizer efficiency, in characters per token. Our monolingual Sámi tokenizer trained on the \textsc{Lapt} data obtains 4.5 characters per token on that data, but this drops to 1.9 and 1.6 on the UD North Sámi and Skolt Sami datasets, respectively; this indicates a significant domain shift between the text seen in pre-training and in the UD datasets. We hypothesize that the model overcomes this vocabulary issue by available fine-tuning data for North Sámi, but that this does not occur for Skolt Sámi, since we evaluate it in a \textit{zero-shot} setting.

In addition, the tokenizer shows a dramatic increase in OOV tokens when applied to Skolt Sámi --- the unigram frequency for \texttt{<unk>} increases to 9\%, from only 0.3\% on the \textsc{Lapt} data.\footnote{North Sámi OOV frequency is only 0.003\%.} Single-character tokens like \texttt{<õ>}, \texttt{<ä>}, \texttt{<â>}, and \texttt{<å>} also greatly increased in frequency, demonstrating the substantial hindrance that orthography differences can have on transfer between otherwise closely-related languages. These findings once again highlight importance of \textit{quality} for language-modeling data, even when large web-scraped datasets have become the norm \cite{kreutzer-etal-2022-quality}. Consequently, a future best practice may be to consider the intended downstream tasks (and their text distributions) when forming the vocabulary for a specialized multilingual model in order to minimize the occurrences of UNK tokens and facilitate better transfer learning between the language-modeling and task domains.


\subsection{Best practices}\label{sec:takeaways}

\paragraph{Multilingualism is beneficial for many languages}
The baselines in \S\ref{sec:baseline_results} demonstrate that given an overall computational budget, it is more effective to adapt a multilingual model to jointly cover a group of languages than it is to adapt models for each individual language. This is especially true for low-resource languages, but surprisingly some high-resource languages like Hungarian and Russian also benefit from multilingual training. This supports the idea that multilingual training is useful for learning general patterns that are beneficial to the performance of many languages. Table~\ref{tab:zeroshot_mono_baseline} further shows that robust performance for low-resource languages like Komi and Moksha, which lack task fine-tuning sets, is only feasible with the combination of multilinguality and transfer learning.

\paragraph{Specialized vocab is more effective and efficient}
Our multilingual baselines in \S\ref{sec:baseline_results} demonstrate that even models with our smallest specialized vocabulary are on par with or outperform those retaining the large ``cross-lingual'' vocabulary from XLM-R, regardless of language. Table~\ref{tab:vocab_and_compute} shows that the 16k vocabulary tokenizes Uralic data with similar efficiency as the XLM-R vocabulary (in terms of mean sequence length), while yielding a model that is 35\% of XLM-R's size. This reduction is significant both for the size of the model in disk/memory and for computational cost during training.\footnote{Per \citet{kaplan_scaling_2020}, we estimate the number of operations per training step, per token as $6(N + dv + 2d)$, where $N$ is the number of non-embedding parameters, $d$ is the hidden dimension, and $v$ is the vocabulary size. Note this estimate is approximately proportional to the total number of parameters.}

\paragraph{Training steps vs. vocabulary size}
\begin{table}[h]
    \centering
    \begin{tabular}{cll}
        \toprule
        Vocab size & Parameters & Avg. length \\
        \midrule
        16k & 98.6M & 49.9 \\
        32k & 111.2M (+13\%) & 44.3 (-11\%) \\
        64k & 136.4M (+23\%) & 39.7 (-10\%) \\
        128k & 186.8M (+37\%) & 36.1 (-9\%) \\
        \midrule
        250k (orig) & 278.3M & 48.4 \\
        \bottomrule
    \end{tabular}
    \caption{Total number of model parameters and average sequence length for each vocabulary size. In parentheses are percent changes from the next-smallest vocabulary. Sequence length is computed on 100k sentences sampled from the \textsc{lapt} set at $\alpha=0.1$.}
    \label{tab:vocab_and_compute}
\end{table}

Our multi-variable regression analysis reveals that though both training steps and vocabulary size positively contribute to downstream performance in task fine-tuned settings, an additional 100k steps is almost three times as effective as adding 16k additional tokens (\S\ref{sec:statistics}). It should be noted that increasing the vocabulary size from 16k to 32k only increases the number of floating point operations during training about 13\% per token (for XLM-R base), while doubling the training steps doubles the number of operations. At the same time, a larger vocabulary reduces the tokenized sequence length, as the sentencepiece model becomes more efficient; shorter sequences lead to reduced computation.

However, as Table~\ref{tab:vocab_and_compute} shows, every doubling of the vocabulary size only reduces the average sequence length about 10\%, so the parameter increase eventually outpaces efficiency from shorter sequences. Extra parameters also increase the model's memory footprint, which might in turn require more gradient accumulation steps to maintain a constant effective batch size on the same hardware; or it might make the model dependent on higher-tier hardware with more memory.

Finally, our regression analysis shows that vocabulary size does not have a significant effect on task performance in the \textit{zero-shot} setting, which covers our lowest-resource languages (see \S\ref{sec:statistics} and Table~\ref{tab:zeroshot_uas_baselines}). A best practice for adaptation to a low-resource language family might thus be to start with a relatively small vocabulary, and increase the size only until the increase in parameters outpaces the decrease in sequence length. Computational budget can then be spent on longer training rather than a larger model. 

\paragraph{Lower alpha is better overall}
A key finding from our analysis is that sampling alpha values during multilingual training does not have a significant effect on task performance in high-resource languages, while low alphas \textit{do} significantly benefit low-resource languages (\S\ref{sec:statistics}).
Our multilingual models thus frequently achieve their best average performance at the lower $\alpha=0.1$, buoyed by the strong performance of low-resource languages.

This finding indicates that practitioners can aggressively up-sample lower-resource languages in multilingual datasets with little risk of degrading the performance of high-resource ``anchor'' languages. Further, as low as $\alpha=0.1$, we see no evidence of ``over-sampling'' these low-resource languages harming downstream performance. However, we note that the high-resource languages we consider are in XLM-R's original pre-training set, which likely affects the model's robustness on those languages. Thus, it is an open question whether the dynamics of multilingual sampling are different in ``from-scratch'' training scenarios or in other high-resource, but previously unseen, languages.

\section{Conclusion}\label{sec:conclusion}
In this work, we show that adapting a pre-trained cross-lingual model to a language family is an effective method for greatly improving NLP task performance for languages in that family, especially those that are under-resourced. Multilingual adaptation soundly outperforms adaptation to single languages for all low-resource Uralic languages we test, as well as for half of the high-resource ones. Further, we show that specializing the model vocabulary for the Uralic family yields significant improvements over models that retain the large ``cross-lingual'' vocabulary of XLM-R, while simultaneously making the model much more computationally efficient and compact in disk/memory. Our statistical analysis of adaptation parameters reveals that both the number of \textsc{lapt} steps and specialized vocabulary size have a significant positive effect on downstream task-finetuned performance. 
However, the language sampling alpha value is only significant for our low-resource languages, indicating that low alpha values can be chosen without significantly affecting high-resource language performance.

We therefore concur with \citet{ogueji-etal-2021-small, ogunremi-etal-2023-mini, chang_when_2023}; \textit{i.a.} that \textit{targeted} or \textit{linguistically-informed} multilingual modeling is one of the most promising avenues for extending NLP advance to the majority of the world's languages. This approach both leverages the benefit of multilingualism for under-resourced languages and avoids the ``Curse of Multilinguality'' seen in massively-multilingual approaches. However, in view of the success of large pre-trained language models, and of the pre-training paradigm more generally \citep{gururangan-etal-2020-dont}, we propose that it is more effective to leverage transferable information in existing cross-lingual models, rather than training targeted models from scratch, as in these previous works. We hope that our findings will inform best practices for such targeted multilingual adaption when extending the benefits of pre-trained models to under-resource languages.

\section*{Limitations}
One limitation of our work is the small selection of evaluation tasks available for under-resourced languages. For most, the only high-quality datasets are found in expertly curated cross-lingual projects such as Universal Dependencies. While a few other datasets exist for under-resourced languages, they are often of questionable quality due to being automatically curated \citep{lignos2022toward}. As such, our experiments are limited to POS tagging and  UAS for dependency parsing.

Second, to maintain a feasible scope of work, we use only XLM-R as a base model for adaptation. Useful future work could include evaluating our adaptation techniques both in larger models, and for ``generative'' models trained with a traditional language modeling task rather than the masked language modeling employed by XLM-R. XGLM \citep{lin-etal-2022-shot}, for example, would be a natural next step, since it is both larger and generative. Evaluating multilingual generative models would also open the door to evaluations on more contemporary prompting-based tasks.

\section*{Acknowledgements}
We thank Ibrahim Sharaf, Anita Silva, and Peter Zuckerman for early investigations of data availability for low-resource languages.

\bibliography{anthology,txlm}

\appendix

\section{Training Details}\label{app:training}
The main details of our experimental process can be found in \S~\ref{sec:experiments}. Here we provide our choice of hyperparameters and other details relevant to reproducibility.

\subsection{Data}
All \textsc{Lapt} data used in our experiments is cleaned and de-duplicated with the OpusFilter package \citep{aulamo-etal-2020-opusfilter}. For low-resource languages, we additionally filter out lines that are identified as English with a probability of 90\% or higher, since positive automatic language-identification for low-resource languages is likely not robust \citep{kreutzer-etal-2022-quality}. We additionally filter out lines composed of less than 2 tokens, lines with an average token length of greater than 16 characters, lines with tokens longer than 32 characters, and lines composed of fewer than 50\% alphabetic characters. We reserve 5\% of the total \textsc{Lapt} data in each language for a development set, and 5\% for a test set.

\subsection{Parameters}
All models are trained and fine-tuned on Nvidia Quadro RTX 6000 GPUs using the Adam optimizer \citep{kingma_adam_2015}. Hyperparameters for Language-Adaptive Pre-Training (\textsc{Lapt}) can be found in Table~\ref{tab:training_hypers}.

\begin{table}[h]
    \centering
    \begin{tabular}{lc}
        \toprule
        Hyperparameter & Value \\
        \midrule
        \texttt{mlm\_masking\_prob} & 0.15 \\
        \texttt{max\_sequence\_length} & 256 \\
        \texttt{learning\_rate} & 1e-5 \\
        \texttt{lr\_schedule} & linear \\
        \texttt{batch\_size} & 200 \\
        \texttt{max\_gradient\_norm} & 1.0 \\
        \bottomrule
    \end{tabular}
    \caption{Hyperparameters for model training (\textsc{Lapt})}
    \label{tab:training_hypers}
\end{table}

\section{Evaluation Details}\label{app:evaluation}

\subsection{Data}
Most language have a standard train/dev/test split curated the original Universal Dependencies dataset \citep{de-marneffe-etal-2021-universal}. Erzya, however, only has a standard train/test split. To form a dev split, we randomly sample 300 sentences from the train split. The inventory of UD evaluation data can be found in Table~\ref{tab:eval_data_inventory}.

\begin{table*}[h]
    \centering
    \begin{tabular}{lcccccc}
        \toprule
        Language & Code & Branch & Script & Train & Dev & Test \\
        \midrule
        Russian & ru & n/a & Cyrillic & 69,630 & 8,906 & 8,800 \\
        Finnish & fi & Finnic & Latin & 14,981 & 1,875 & 1,867 \\
        Estonian & et & Finnic & Latin & 5,444 & 833 & 913 \\
        North Sámi & sme & Sámi & Latin & 2,001 & 256 & 865 \\
        Hungarian & hu & Hungarian & Latin & 910 & 441 & 449 \\
        Erzya & myv & Mordvinic & Cyrillic & 896 & 300 & 921 \\
        Komi & koi & Permic & Cyrillic & 0 & 0 & 663 \\
        Moksha & mdf & Mordvinic & Cyrillic & 0 & 0 & 446 \\
        Skolt Sámi & sms & Sámi & Latin & 0 & 0 & 244 \\
        Karelian & krl & Finnic & Latin & 0 & 0 & 228 \\
        Livvi & olo & Finnic & Latin & 0 & 0 & 106 \\
        \bottomrule
    \end{tabular}
    \caption{Universal Dependencies evaluation set sizes, by number of examples (sentences)}
    \label{tab:eval_data_inventory}
\end{table*}

\subsection{Parameters}
Hyperparameters for task fine-tuning on POS and UAS are in Table~\ref{tab:eval_hypers}. We cap fine-tuning training data at 32,768 sequences (only relevant for Russian).

\begin{table}[h]
    \centering
    \begin{tabular}{lc}
        \toprule
        Hyperparameter & Value \\
        \midrule
        \texttt{max\_sequence\_length} & 256 \\
        \texttt{learning\_rate} & 5e-6 \\
        \texttt{lr\_schedule} & constant \\
        \texttt{max\_epochs} & 64 \\
        \texttt{eval\_interval} (epochs) & 2 \\
        \texttt{patience} (epochs) & none / 8 \\
        \texttt{batch\_size} & 72 \\
        \texttt{max\_gradient\_norm} & 1.0 \\
        \bottomrule
    \end{tabular}
    \caption{Hyperparameters for model task fine-tuning. \textit{few-shot} has no early stopping. \textit{Full-finetune} and \textit{zero-shot} settings have early stopping after patience of 8 epochs}
    \label{tab:eval_hypers}
\end{table}

\subsection{Unlabeled Attachment Score}
Unlabeled Attachment Score (UAS) is the accuracy with which a model assigns each word its proper dependency head. Our implementation uses the graph biaffine algorithm defined in \citet{dozat_deep_2017}. The contextual embedding representation for each token $r_i$ is passed through each of two feed-forward layers, to produce a representation of this token as a head and as a dependent, respectively:
    \[ h_i^{head} = \text{FFN}^{head}(r_i)\]
    \[ h_i^{dep} = \text{FFN}^{dep}(r_i)\]

The score of a directed edge i $\rightarrow$ j, is then assigned according to a biaffine scoring function:
\[\text{Biaffine}(h_i^{head},h_j^{dep}) = U_{\text{arc}} + W_{\text{arc}} + b\]
\[U_{\text{arc}} = h_j^{dep} \cdot U_{\text{arc\_head}}^T \]
\[U_{\text{arc\_head}} = U \cdot h_i^{head}\]
\[W_{\text{arc}} = W \cdot h_i^{head} \]

where U, W, and b are weights learned by the model. A probability distribution over possible heads is then computed by passing score(i → j) through a softmax layer. Our implementation is based on \citet{jurafsky_speech_2024} and \url{https://www.cse.chalmers.se/~richajo/nlp2019/l7/Biaffine%20dependency%20parsing.html}.

\section{Additional results}\label{app:results}

\begin{figure*}[h]
    \begin{center}
    \includegraphics[width=\textwidth]{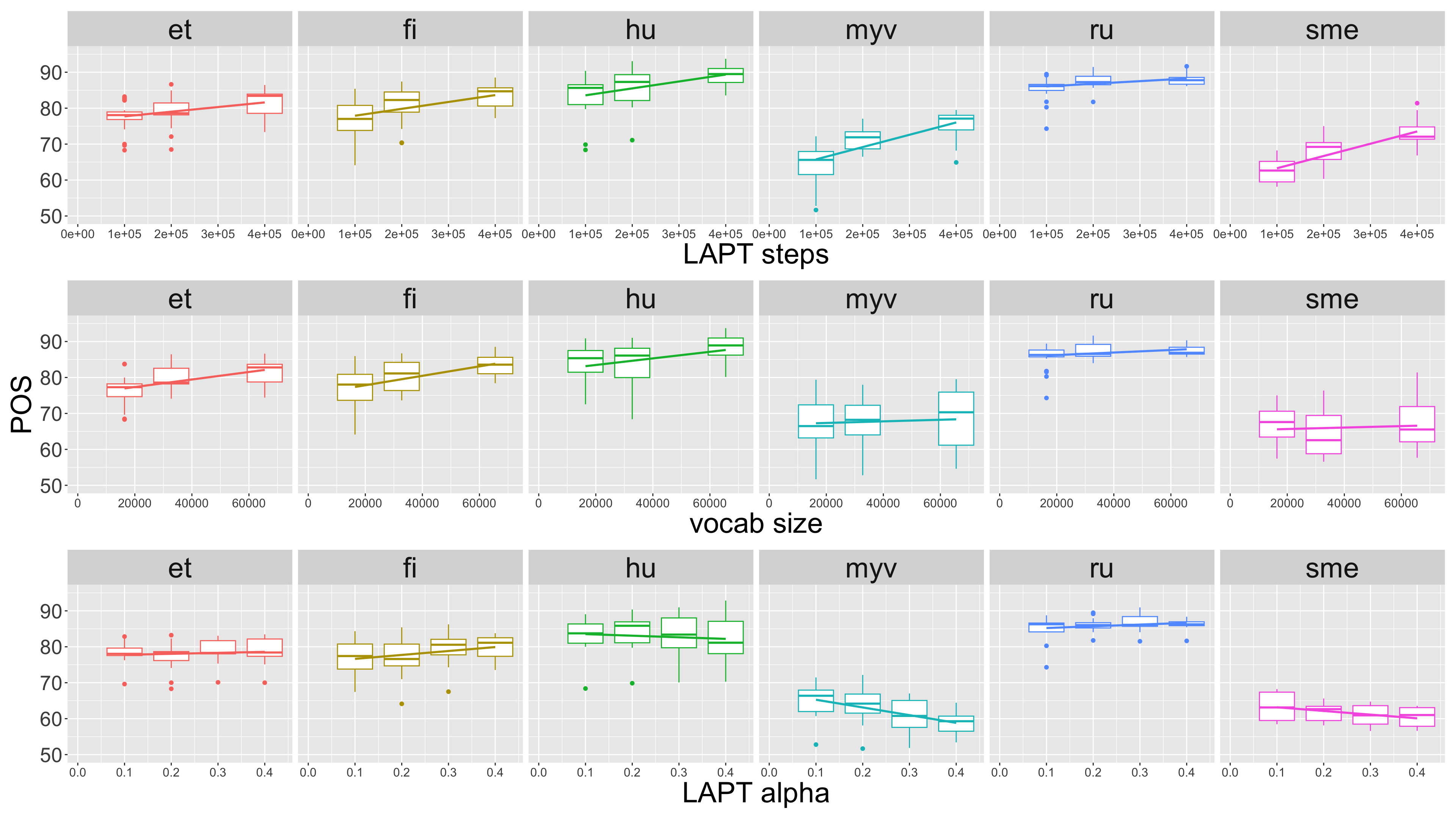}
    \caption{Few-shot POS --- effect of hyper-parameters by language, marginalized across other parameter settings}
    \label{fig:fewshot_pos_plots}
    \end{center}
\end{figure*}

\begin{figure*}[h]
    \begin{center}
    \includegraphics[width=0.85\textwidth]{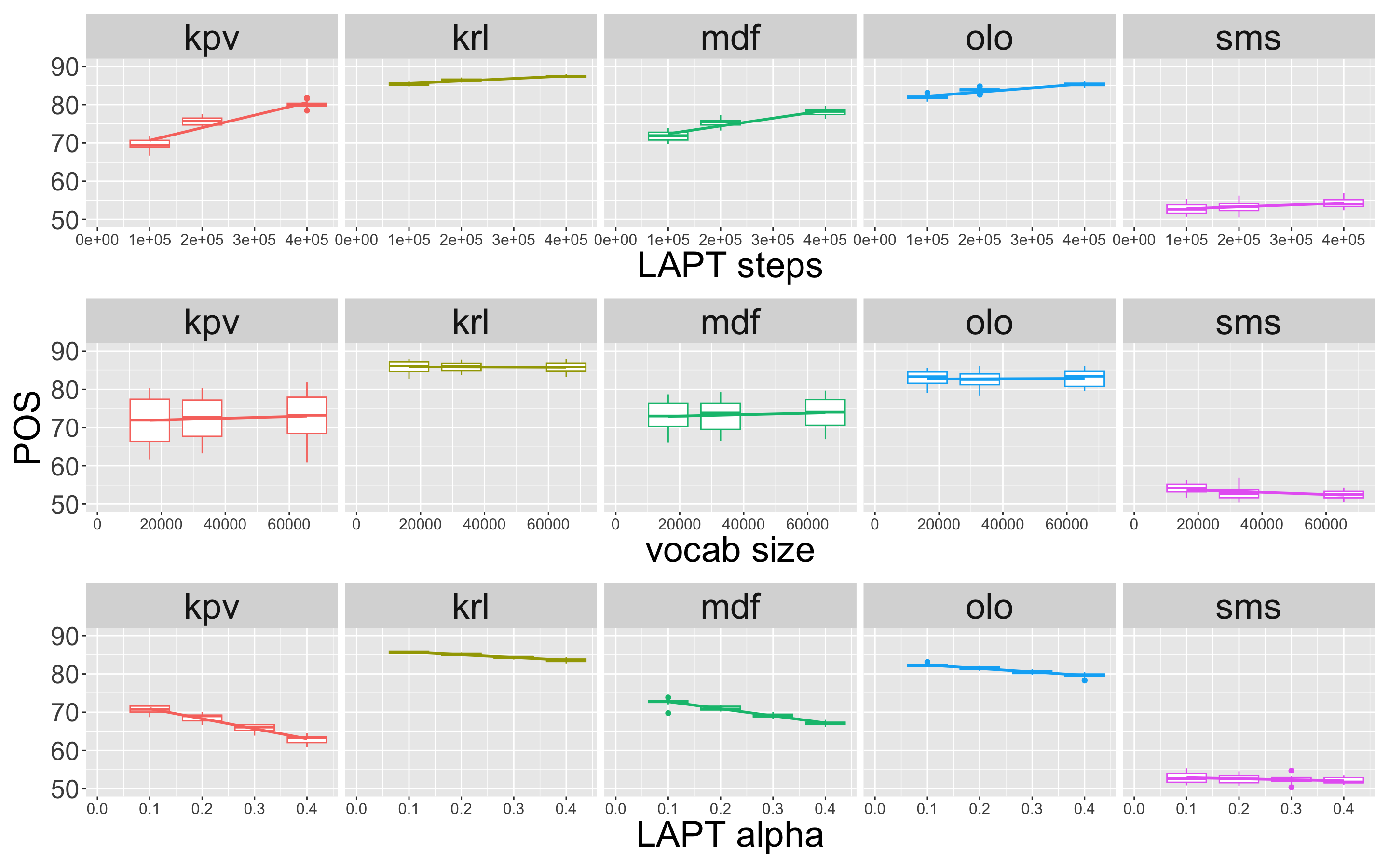}
    \caption{Zero-shot POS --- effect of hyper-parameters by language, marginalized across other parameter settings}
    \label{fig:zeroshot_pos_plots}
    \end{center}
\end{figure*}

Results and visualizations for the POS task can be found in this appendix. For POS, the multilingual baseline without vocabulary specialization performs more on-par with models with specialized vocabulary (Tables~\ref{tab:fewshot_pos_baselines}, \ref{tab:zeroshot_pos_baselines}). This is possibly due to the relative simplicity of the task. The parameter-wise trends for POS are mostly the same as for UAS (Figures~\ref{fig:fewshot_pos_plots}, \ref{fig:zeroshot_pos_plots}).

\begin{table*}[h]
    \centering
    \resizebox{\textwidth}{!}{
    \begin{tabular}{cccccccccc}
        \toprule
        \textsc{Lapt} & Alpha & Vocab & Erzya & North Sámi & Estonian & Finnish & Hungarian & Russian & Avg \\
        \midrule
        0 & * & 250k (orig) & 50.9 $\pm$ 1.9 & 53.8 $\pm$ 3.1 & 63.9 $\pm$ 5.4 & 66.7 $\pm$ 3.7 & 81.5 $\pm$ 5.4 & 86.8 $\pm$ 1.0 & 67.3 \\
        400k & 0.1 & 250k (orig) & 75.2 $\pm$ 2.6 & \textbf{77.2 $\pm$ 2.6} & \textbf{84.2 $\pm$ 0.3} & \textbf{83.3 $\pm$ 2.1} & 88.0 $\pm$ 3.2 & \textbf{90.1 $\pm$ 2.0} & 82.7 \\
        \midrule
        400k & 0.1 & 16k & 76.1 $\pm$ 3.3 & 73.2 $\pm$ 1.2 & 77.7 $\pm$ 3.9 & 79.7 $\pm$ 2.6 & 89.3 $\pm$ 1.3 & 87.5 $\pm$ 0.5 & 80.6 \\
        400k & 0.1 & 32k & 72.3 $\pm$ 4.2 & 71.4 $\pm$ 1.2 & 82.7 $\pm$ 2.4 & 82.3 $\pm$ 3.8 & 87.7 $\pm$ 2.4 & 88.0 $\pm$ 2.2 & 80.7 \\
        400k & 0.1 & 64k & \textbf{78.0 $\pm$ 1.4} & \textbf{76.5 $\pm$ 3.5} & 83.0 $\pm$ 2.4 & \textbf{85.4 $\pm$ 2.2} & \textbf{94.1 $\pm$ 1.1} & 88.1 $\pm$ 1.5 & \textbf{84.2} \\
        \bottomrule
    \end{tabular}}
    \caption{Few-shot POS --- comparison with multilingual baselines. First row is XLM-R ``off-the-shelf'' (without \textsc{Lapt} or vocabulary replacement). Second row is XLM-R with original cross-lingual vocabulary, but fine-tuned on Uralic languages with \textsc{Lapt}}
    \label{tab:fewshot_pos_baselines}
\end{table*}

\begin{table*}[h]
    \centering
    \resizebox{\textwidth}{!}{
    \begin{tabular}{ccccccccc}
        \toprule
        \textsc{Lapt} & Alpha & Vocab & Karelian & Komi & Livvi & Moksha & Skolt Sámi & Avg \\
        \midrule
        0 & * & 250k (orig) & 77.7 $\pm$ 0.6 & 49.6 $\pm$ 0.6 & 73.7 $\pm$ 0.8 & 64.4 $\pm$ 0.3 & 55.0 $\pm$ 1.2 & 64.1 \\
        400k & 0.1 & 250k (orig) & 86.7 $\pm$ 0.2 & 80.0 $\pm$ 0.2 & 85.2 $\pm$ 0.4 & 79.4 $\pm$ 0.2 & \textbf{56.1 $\pm$ 1.0} & \textbf{77.5} \\
        \midrule
        400k & 0.1 & 16k & \textbf{87.7 $\pm$ 0.2} & 80.0 $\pm$ 0.3 & 85.0 $\pm$ 0.2 & 78.3 $\pm$ 0.2 & \textbf{55.4 $\pm$ 0.3} & 77.3 \\
        400k & 0.1 & 32k & 87.3 $\pm$ 0.3 & 80.1 $\pm$ 0.2 & \textbf{85.6 $\pm$ 0.4} & 78.6 $\pm$ 0.5 & 53.7 $\pm$ 0.3 & 77.0 \\
        400k & 0.1 & 64k & 87.4 $\pm$ 0.4 & \textbf{81.4 $\pm$ 0.4} & \textbf{85.6 $\pm$ 0.2} & \textbf{79.6 $\pm$ 0.1} & 52.2 $\pm$ 1.7 & 77.2 \\
        \bottomrule
    \end{tabular}}
    \caption{Zero-shot POS --- comparison with multilingual baselines. First row is XLM-R ``off-the-shelf'' (without \textsc{Lapt} or vocabulary replacement). Second row is XLM-R with original cross-lingual vocabulary, but fine-tuned on Uralic languages with \textsc{Lapt}}
    \label{tab:zeroshot_pos_baselines}
\end{table*}

\section{Regression tables}\label{app:regressions}
The full regression summaries from the \texttt{lme4} package \citep{bates_fitting_2015} can be found in Tables~\ref{tab:finetuned_regression_a}-\ref{tab:zeroshot_regression_b}. These cover both the fine-tuned (\textit{few-shot}/\textit{full-finetune}) and \textit{zero-shot} models. As mentioned in \S~\ref{sec:experiments}, we test four values of alpha for experiments with 100k steps, but only two values for longer experiments. Because this introduces artificial correlation of input variables, we separate the regression with two alphas as our ``main'' results, but include the summary of regressions with four values (but no variation in training steps) here (Tables~\ref{tab:finetuned_regression_b} and \ref{tab:zeroshot_regression_b}). These secondary regressions show a greater effect size for low-resource alpha, indicating the estimate between the alpha values 0.1 and 0.2 might not accurate estimate the larger trends. Note that these secondary regressions do not change the standings of which variables are significant.

\begin{table*}[h]
    \centering
    \begin{tabular}{lrrrrr}
        \toprule
        Fixed effects & Estimate & Std. Errror & df & t value & p value \\
        \midrule
        \texttt{(Intercept)} & \textbf{75.93} & 2.53 & 5.63 & 29.97 & \textbf{2.00e-07} \\
        \texttt{lapt\_steps} & \textbf{1.67} & 0.15 & 1691.67 & 11.16 & \textbf{< 2e-16} \\
        \texttt{vocab\_size} & \textbf{0.62} & 0.15 & 1691.67 & 4.15 & \textbf{3.49e-05} \\
        \texttt{finetuning\_lines} & \textbf{0.40} & 0.01 & 1696.77 & 30.32 & \textbf{< 2e-16} \\
        \texttt{taskuas} & \textbf{-13.84} & 0.38 & 1691.67 & -36.71 & \textbf{< 2e-16} \\
        \texttt{resourcehigh:lapt\_alpha} & 0.42 & 0.46 & 1582.98 & 0.92 & 0.3606 \\
        \texttt{resourcelow:lapt\_alpha} & \textbf{-1.36} & 0.64 & 1239.05 & -2.11 & \textbf{0.0347} \\
        \bottomrule
    \end{tabular}
    \caption{Regression summary table for \textit{few-shot} and \textit{full-finetune} settings. Significant coefficients and p values in bold. This regression covers all training lengths (step numbers), but only includes alphas \{0.1, 0.2\}. Formula:\\ \small{\texttt{lmer(accuracy $\sim$ lapt\_steps + vocab\_size + finetuning\_lines + task + resource:lapt\_alpha + (1 | language))}}}
    \label{tab:finetuned_regression_a}
\end{table*}

\begin{table*}[h]
    \centering
    \begin{tabular}{lrrrrr}
        \toprule
        Fixed effects & Estimate & Std. Errror & df & t value & p value \\
        \midrule
        \texttt{(Intercept)} & \textbf{78.39} & 2.95 & 5.39 & 26.61 & \textbf{6.27e-07} \\
        \texttt{vocab\_size} & \textbf{0.39} & 0.19 & 1140.76 & 2.01 & \textbf{0.0448} \\
        \texttt{finetuning\_lines} & \textbf{0.42} & 0.02 & 1146.00 & 25.14 & \textbf{< 2e-16} \\
        \texttt{taskuas} & \textbf{-14.16} & 0.48 & 1140.76 & -29.44 & \textbf{< 2e-16} \\
        \texttt{resourcehigh:lapt\_alpha} & 0.19 & 0.26 & 1132.87 & 0.72 & 0.4730 \\
        \texttt{resourcelow:lapt\_alpha} & \textbf{-2.38} & 0.37 & 1058.70 & -6.45 & \textbf{1.66e-10} \\
        \bottomrule
    \end{tabular}
    \caption{Secondary regression summary table for \textit{few-shot} and \textit{full-finetune} settings. Significant coefficients and p values in bold. This regression covers all values of alpha \{0.1, 0.2, 0.3, 0.4\}, which are only tested in experiments with 100k training steps. Thus, the \texttt{lapt\_steps} variable is excluded from this regression. Formula:\\ \texttt{lmer(accuracy $\sim$ vocab\_size + finetuning\_lines + task + resource:lapt\_alpha + (1 | language))}}
    \label{tab:finetuned_regression_b}
\end{table*}

\begin{table*}[h]
    \centering
    \begin{tabular}{lrrrrr}
        \toprule
        Fixed effects & Estimate & Std. Errror & df & t value & p value \\
        \midrule
        \texttt{(Intercept)} & \textbf{72.68} & 5.20 & 4.09 & 13.99 & \textbf{1.31e-4} \\
        \texttt{lapt\_steps} & \textbf{1.35} & 0.11 & 711.00 & 12.58 & \textbf{< 2e-16} \\
        \texttt{vocab\_size} & 0.04 & 0.11 & 711.00 & 0.35 & 0.7266 \\
        \texttt{lapt\_alpha} & \textbf{-0.81} & 0.27 & 711.00 & -3.02 & \textbf{2.66e-3} \\
        \texttt{taskuas} & \textbf{-12.89} & 0.27 & 711.00 & -48.02 & \textbf{< 2e-16} \\
        \bottomrule
    \end{tabular}
    \caption{Regression summary table for \textit{zero-shot} setting. Significant coefficients and p values in bold. This regression covers all training lengths (step numbers), but only includes alphas \{0.1, 0.2\}. Formula:\\ \texttt{lmer(accuracy $\sim$ lapt\_steps + vocab\_size + lapt\_alpha + task + (1 | language))}}
    \label{tab:zeroshot_regression_a}
\end{table*}

\begin{table*}[h]
    \centering
    \begin{tabular}{lrrrrr}
        \toprule
        Fixed effects & Estimate & Std. Errror & df & t value & p value \\
        \midrule
        \texttt{(Intercept)} & \textbf{74.33} & 4.72 & 4.08 & 15.73 & \textbf{8.31e-5} \\
        \texttt{vocab\_size} & -0.05 & 0.12 & 472.00 & -0.38 & 0.7020 \\
        \texttt{lapt\_alpha} & \textbf{-1.30} & 0.14 & 472.00 & -9.46 & \textbf{< 2e-16} \\
        \texttt{taskuas} & \textbf{-12.45} & 0.31 & 472.00 & -40.55 & \textbf{< 2e-16} \\
        \bottomrule
    \end{tabular}
    \caption{Secondary regression summary table for \textit{zero-shot} setting. Significant coefficients and p values in bold. This regression covers all values of alpha \{0.1, 0.2, 0.3, 0.4\}, which are only tested in experiments with 100k training steps. Thus, the \texttt{lapt\_steps} variable is excluded from this regression. Formula:\\ \texttt{lmer(accuracy $\sim$ vocab\_size + lapt\_alpha + task + (1 | language))}}
    \label{tab:zeroshot_regression_b}
\end{table*}

\end{document}